\documentclass[sigconf]{acmart}

\usepackage{booktabs}
\usepackage{amsmath}
\usepackage{graphicx}
\usepackage{xcolor}
\usepackage{multirow}
\usepackage{tikz}
\usetikzlibrary{arrows.meta, positioning}

\copyrightyear{2026}
\acmYear{2026}
\setcopyright{cc}
\setcctype{by}
\acmConference[MM '26]{Proceedings of the 34th ACM International Conference on Multimedia}{November 10--14, 2026}{Rio de Janeiro, Brazil}
\acmBooktitle{Proceedings of the 34th ACM International Conference on Multimedia (MM '26), November 10--14, 2026, Rio de Janeiro, Brazil}
\acmDOI{10.1145/3767308.3836082}
\acmISBN{979-8-4007-2213-4/2026/11}

\begin{document}

\title{Interpreting Video Representations with\\Spatio-Temporal Sparse Autoencoders}

\author{Atahan Dokme}
\affiliation{
  \institution{Georgia Institute of Technology}
  \city{Atlanta}
  \state{Georgia}
  \country{USA}
}
\email{adokme3@gatech.edu}
\orcid{0009-0001-0483-2395}

\author{Sriram Vishwanath}
\affiliation{
  \institution{Georgia Institute of Technology}
  \city{Atlanta}
  \state{Georgia}
  \country{USA}
}
\email{sriram@ece.gatech.edu}
\orcid{0000-0003-3112-4885}

\begin{abstract}
We present the first systematic study of Sparse Autoencoders (SAEs) on video representations. Standard SAEs decompose video into 
interpretable, monosemantic features but destroy temporal coherence: hard TopK selection produces unstable feature assignments across 
frames, reducing autocorrelation by 36\%. We propose spatio-temporal contrastive objectives and Matryoshka hierarchical grouping that 
recover and even exceed raw temporal coherence. The contrastive loss weight controls a tunable trade-off between reconstruction and temporal coherence. A systematic ablation on two backbones and two datasets shows that different configurations excel at different goals: reconstruction fidelity, temporal coherence, action discrimination, or interpretability. Contrastive SAE features improve action classification by $+3.9$ pp over raw features and text-video retrieval by up to 2.8$\times$ R@1. A cross-backbone analysis reveals that standard monosemanticity metrics contain a backbone-alignment artifact: both DINOv2 and VideoMAE produce equally monosemantic features under an independent (CLIP) similarity space. Targeted feature ablation shows that contrastive training concentrates the probe's predictive signal into a small number of identifiable features. Supplementary material, code, configurations and evaluation scripts are available at \url{https://github.com/atahandokme/spatio-temporal-sparse-autoencoders-video}.
\end{abstract}



\maketitle

\section{Introduction}
\label{sec:intro}

Sparse Autoencoders (SAEs) have transformed the interpretability of language models~\cite{bricken2023monosemanticity, templeton2024scaling} and, more recently, static image encoders~\cite{pach2025sparse, stevens2025interpretabletestablevisionfeatures}. Yet video vision models remain underexplored. From self-supervised encoders like DINOv2~\cite{oquab2024dinov} and VideoMAE~\cite{tong2022videomaemaskedautoencodersdataefficient} to multimodal retrieval and captioning systems, we lack tools to understand what these models learn internally. The gap is practically pressing: multimodal assistants and systems that operate graphical interfaces consume video and screen streams through frozen encoders of exactly this kind, and SAE interventions on a vision encoder are known to steer the multimodal LLM above it~\cite{pach2025sparse} --- which, for video, requires features that stay attached to a concept as the scene evolves. Whether SAEs can meaningfully decompose \textit{video} representations, where features are spatio-temporal rather than spatial, is an open question.

We observe that the challenge is structural. A video clip produces a feature tensor spanning $T$ frames and $P$ spatial patches, where the same semantic concept (a hand, a moving object) should ideally activate consistent features across time. Standard SAEs, encoding each patch independently, are oblivious to this structure. This raises two questions: \textit{Can SAEs reveal interpretable structure in video features?} And \textit{does the temporal dimension require new methodology?}

We find that standard SAEs produce monosemantic video features but fail to preserve temporal coherence, making it impossible to track concepts across frames. Our central finding is a design principle: the contrastive strength controls the trade-off between reconstruction fidelity and temporal coherence. Both the loss coefficient $\lambda$ and the temperature $\tau$ modulate this strength through different mechanisms ($\lambda$ scales the loss magnitude, $\tau$ controls the sharpness of positive/negative separation), but both move variants along the same Pareto frontier. This is not a limitation to work around. It is a controllable axis that practitioners can set based on their application. Our contributions:

\begin{enumerate}
    \item \textbf{Systematic empirical study of SAEs on video representations.} Our ``first systematic study'' claim is scoped specifically to \textit{video} representations; the SAE-for-vision framing, the MonoSemanticity metric, and the evaluation protocol we use are due to Pach et al.~\cite{pach2025sparse}, and we make no novelty claim over SAEs for vision in general. We train TopK SAEs on two backbones (DINOv2 and VideoMAE) and two datasets (SSv2 and K400), demonstrating that SAEs decompose video features into monosemantic, interpretable concepts such as scene elements, objects, and action-correlated features. Video-native models (VideoMAE) score substantially higher on own-space monosemanticity metrics than image models (MS $\approx 0.97$ vs.\ $\approx 0.36$), but a CLIP-based control reveals both produce equally monosemantic features (MS $\approx$ 0.68), identifying a metric alignment artifact in the standard evaluation protocol.

    \item \textbf{Diagnosis of temporal incoherence as a core failure mode.} We show that standard SAEs destroy temporal structure: lag-1 autocorrelation drops 36\%, with 95\% of features below the 0.3 coherence threshold. We trace this to the hard top-$k$ selection mechanism and show via targeted feature ablation that the resulting features, while temporally unstable, are nonetheless predictively important for action discrimination: zeroing the ten highest-weighted features of the temporal contrastive SAE costs $9.3$ pp of accuracy, whereas zeroing ten random features costs nothing.

    \item \textbf{A controllable reconstruction--coherence trade-off.} Extending T-SAE~\cite{bhalla2026temporal} from text to video, we evaluate three contrastive variants (temporal, separate, raster-scan) and Matryoshka hierarchical grouping. A systematic ablation across 29 configurations reveals that the contrastive strength (via $\lambda$ or $\tau$) controls where a variant sits on the Pareto frontier. At low weight ($\lambda{=}0.01$), all variants preserve reconstruction ($R^2 = 0.31{-}0.42$). At high weight, Temporal+M exceeds raw temporal coherence (lag-1 $= 0.462$ vs.\ 0.435). The best configurations improve action discrimination by $+3.9$ pp over raw features and boost text-video retrieval by 2.8$\times$ R@1 on DINOv2. Matryoshka grouping concentrates action-relevant information into 20\% of the dictionary.
\end{enumerate}

\section{Related Work}
\label{sec:related}

\paragraph{Sparse Autoencoders for Interpretability.}
SAEs were introduced to decompose neural network activations into sparse, monosemantic features~\cite{bricken2023monosemanticity, cunningham2023sparseautoencodershighlyinterpretable}. Subsequent work scaled SAEs to frontier language models~\cite{templeton2024scaling}. Key architectural innovations include TopK activation~\cite{gao2024scalingevaluatingsparseautoencoders}, JumpReLU~\cite{rajamanoharan2024jumpingaheadimprovingreconstruction}, BatchTopK~\cite{bussmann2024batchtopksparseautoencoders} and Matryoshka SAEs~\cite{bussmann2025matryoshka} for hierarchical feature organization, and auxiliary losses for dead latent recovery.

\paragraph{SAEs for Vision Models.}
Recent work has extended SAEs to vision encoders, establishing that sparse decomposition yields interpretable visual features. Pach et al.~\cite{pach2025sparse} train SAEs on CLIP and introduce the MonoSemanticity evaluation framework we adopt in this work, demonstrating that SAE interventions on the vision encoder can steer multimodal LLM outputs. Stevens et al.~\cite{stevens2025interpretabletestablevisionfeatures} enable patch-level causal edits via SAEs on frozen ViTs, providing a rigorous causal methodology that motivates our ablation experiments. Joseph et al.~\cite{joseph2025steeringclipsvisiontransformer} provide a steerability analysis of CLIP's vision transformer, and Lim et al.~\cite{lim2025sparse} introduce PatchSAE for concept-level analysis during adaptation. While these works demonstrate the promise of SAEs for vision, they all operate on static images. None examine how the temporal structure of video interacts with sparse decomposition, which is the focus of our study.

\paragraph{Temporal SAEs.}
Bhalla et al.~\cite{bhalla2026temporal} propose Temporal SAEs (T-SAEs) for language models, adding contrastive losses on adjacent tokens to disentangle semantic from positional features. They demonstrate that temporal consistency improves feature quality for text sequences. We extend this temporal contrastive principle from 1D text sequences to spatio-temporal video representations, where the challenge is qualitatively different: video tokens have both spatial adjacency within frames and temporal adjacency across frames, and the visual content at a given spatial position can change substantially between frames due to motion. Our raster-scan variant (Section~\ref{sec:method_st}) serializes both axes into a single contrastive objective, though its cross-frame coupling arises only at frame boundaries rather than at matched spatial positions.

\paragraph{Temporal Consistency in Video Representations.}
Object-centric approaches such as Slot Contrast~\cite{manasyan2025temporallyconsistentobjectcentriclearning} learn temporally consistent slots by contrasting object representations across frames. Spatio-temporal attention attribution methods~\cite{wang2024staaspatiotemporalattentionattribution} provide post-hoc explanations of video model predictions. Temporal contrastive self-supervised methods such as TCLR~\cite{Dave_2022} promote temporal diversity, while multi-pretext frameworks like CSTP~\cite{zhang2021contrastivespatiotemporalpretextlearning} learn temporal structure during pretraining. Optical flow methods like RAFT~\cite{teed2020raftrecurrentallpairsfield} provide dense correspondences that could serve as temporal pairing targets. Our approach is complementary to these methods. We decompose frozen features into sparse dictionaries rather than learning new representations or modifying pretraining. Our adjacency-based contrastive losses operate on fixed spatial positions rather than tracked objects. This is a deliberate design choice: SAE features should decompose the model's representation as-is, without introducing external correspondence mechanisms. We discuss the limitations of fixed spatial pairing under camera motion in Section~\ref{sec:discussion}.

\paragraph{Interpretability for Action Models.}
Häon et al.~\cite{haon2025mechanisticinterpretabilitysteeringvisionlanguageaction} apply mechanistic interpretability to Vision-Language-Action models for robotics using logit lens projections. Khan et al.~\cite{khan2025controlling} use SAEs on VLA residual streams to construct steering vectors for robotic control. Both operate in the robotics domain with action-specific models. Our work addresses general-purpose video vision models and establishes interpretability foundations applicable to any video understanding system.

\section{Method}
\label{sec:method}

\subsection{Applying SAEs to Video Features}
\label{sec:method_video}

Given a video clip, we extract intermediate features from a vision backbone, obtaining a tensor $\mathbf{X} \in \mathbb{R}^{T \times P \times D}$ with $T$ frames, $P$ spatial patches per frame, and $D$-dimensional features. For DINOv2 ViT-B/14, we obtain $(T{=}16, P{=}256, D{=}768)$; for VideoMAE-B, $(T{=}8, P{=}196, D{=}768)$.

To train a standard SAE, we flatten $\mathbf{X}$ to $(T \cdot P, D)$ and treat each patch as an independent token. Each such token is written $\mathbf{x} = \mathbf{X}_{t,p,:} \in \mathbb{R}^{D}$, the feature at spatial patch $p$ of frame $t$. A TopK SAE~\cite{gao2024scalingevaluatingsparseautoencoders} learns an overcomplete representation:
\begin{align}
    \mathbf{z} &= \text{TopK}(\text{ReLU}(W_e (\mathbf{x} - \mathbf{b}_{\text{pre}}) + \mathbf{b}_e), k) \\
    \hat{\mathbf{x}} &= W_d \mathbf{z} + \mathbf{b}_{\text{pre}}
\end{align}
where $H \gg D$ is the dictionary size, $\text{TopK}(\cdot, k)$ retains only the $k$ largest activations, and the loss combines reconstruction with auxiliary dead latent recovery:
\begin{equation}
    \mathcal{L}_{\text{base}} = \|\mathbf{x} - \hat{\mathbf{x}}\|_2^2 + \alpha \mathcal{L}_{\text{aux}}
\end{equation}

Here $\mathbf{z} \in \mathbb{R}^{H}$ is the sparse activation vector, $W_e \in \mathbb{R}^{H \times D}$ and $W_d \in \mathbb{R}^{D \times H}$ are the encoder and decoder weight matrices, $\mathbf{b}_e \in \mathbb{R}^{H}$ is the encoder bias, and $\mathbf{b}_{\text{pre}} \in \mathbb{R}^{D}$ is the pre-encoder (input) bias.

This approach is agnostic to the spatio-temporal structure of video. The SAE sees no difference between patches from the same frame and patches from different frames.

\subsection{Spatio-Temporal SAE Variants}
\label{sec:method_st}

To preserve temporal and spatial structure, we propose three contrastive variants and a hierarchical grouping mechanism. All contrastive variants augment $\mathcal{L}_{\text{base}}$ with InfoNCE losses~\cite{oord2019representationlearningcontrastivepredictive} that encourage consistent SAE activations between nearby patches, using negative pairs to prevent collapse toward trivially uniform features.

\paragraph{Variant 1: Temporal Contrastive (T-SAE)}
For the same spatial position $p$ across consecutive frames $t$ and $t{+}1$, we encourage consistent activations:
\begin{equation}
    \mathcal{L}_{\text{temp}} = -\frac{1}{N}\sum_{i=1}^{N} \log \frac{\exp(\text{sim}(\mathbf{z}_{t,p}^{(i)}, \mathbf{z}_{t+1,p}^{(i)}) / \tau)}{\sum_{j=1}^{N} \exp(\text{sim}(\mathbf{z}_{t,p}^{(i)}, \mathbf{z}_{t+1,j}^{(i)}) / \tau)}
\end{equation}
where $\text{sim}(\cdot, \cdot)$ denotes cosine similarity and $\tau$ is a temperature parameter. Similarity is computed on the full sparse vectors $\mathbf{z} \in \mathbb{R}^H$ including zeros; since TopK yields exactly $k{=}64$ nonzero entries per token, it is well-defined and nondegenerate. TopK selection is non-differentiable, but the selected values retain their gradients (straight-through on values, hard mask on indices) following Gao et al.~\cite{gao2024scalingevaluatingsparseautoencoders}; we observed no resulting instability. Negatives are all other patches in the batch, from both the same and other clips, giving a mix of hard and easy negatives. This directly extends T-SAE~\cite{bhalla2026temporal} from 1D text sequences to the temporal axis of video.

\paragraph{Variant 2: Separate Contrastive (ST-SAE)}
Video features have structure along \textit{both} time and space. Variant~1 only enforces temporal consistency; ST-SAE adds a spatial contrastive loss between horizontally adjacent patches within each frame. This spatial term takes the same InfoNCE form as $\mathcal{L}_{\text{temp}}$, but pairs horizontally adjacent patches $p$ and $p{+}1$ within the same frame $t$:
\begin{equation}
    \mathcal{L}_{\text{spat}} = -\frac{1}{N}\sum_{i=1}^{N} \log \frac{\exp(\text{sim}(\mathbf{z}_{t,p}^{(i)}, \mathbf{z}_{t,p+1}^{(i)}) / \tau)}{\sum_{j=1}^{N} \exp(\text{sim}(\mathbf{z}_{t,p}^{(i)}, \mathbf{z}_{t,j}^{(i)}) / \tau)}
\end{equation}
Positives are within-frame horizontal neighbours; negatives are all other patches in the batch, exactly as in $\mathcal{L}_{\text{temp}}$. The full objective combines the two:
\begin{equation}
    \mathcal{L} = \mathcal{L}_{\text{base}} + \lambda_t \mathcal{L}_{\text{temp}} + \lambda_s \mathcal{L}_{\text{spat}}
\end{equation}
The two losses operate independently, allowing separate control over temporal and spatial coherence via $\lambda_t$ and $\lambda_s$.

\paragraph{Variant 3: Raster-Scan Contrastive (R-SAE)}
Instead of separate spatial and temporal losses, we serialize the $(T, P)$ grid into a 1D sequence via raster-scan ordering (row-by-row within each frame, then across frames) and apply a single contrastive loss on adjacent elements. Most pairs are within-frame spatial neighbours. At each frame boundary the pair is $(t, P{-}1)$ and $(t{+}1, 0)$: these are \textit{cross-frame boundary pairs} rather than same-position temporal correspondences, so R-SAE introduces a lightweight cross-frame coupling within a single serialized objective, but it does not enforce temporal correspondence in the sense that $\mathcal{L}_{\text{temp}}$ does.

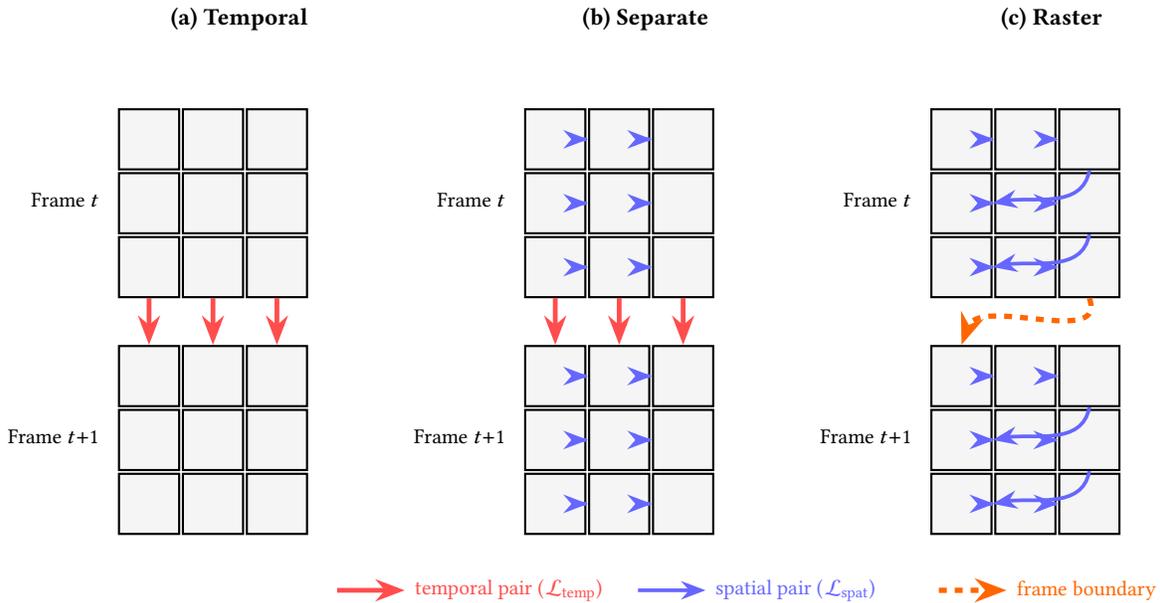
\begin{figure*}[t]
\centering
\begin{tikzpicture}[
    cell/.style={minimum size=0.8cm, draw, thick, inner sep=0pt, font=\small},
    patch/.style={cell, fill=gray!8},
    >=Stealth,
    scale=0.85, transform shape
]
\node[font=\normalsize\bfseries] at (1.2, 1.6) {(a) Temporal};
\node[font=\small, anchor=east] at (-0.55, -0.8) {Frame $t$};
\foreach \r/\ri in {0/0, -0.85/1, -1.7/2} {
  \foreach \c/\ci in {0/0, 0.85/1, 1.7/2} {
    \node[patch] (a1_\ri_\ci) at (\c, \r) {};
  }
}
\node[font=\small, anchor=east] at (-0.55, -3.95) {Frame $t{+}1$};
\foreach \r/\ri in {-3.15/0, -4.0/1, -4.85/2} {
  \foreach \c/\ci in {0/0, 0.85/1, 1.7/2} {
    \node[patch] (a2_\ri_\ci) at (\c, \r) {};
  }
}
\foreach \ci in {0, 1, 2} {
  \draw[->, red!70, line width=1.8pt] (a1_2_\ci.south) -- (a2_0_\ci.north);
}

\node[font=\normalsize\bfseries] at (6.6, 1.6) {(b) Separate};
\node[font=\small, anchor=east] at (4.85, -0.8) {Frame $t$};
\foreach \r/\ri in {0/0, -0.85/1, -1.7/2} {
  \foreach \c/\ci in {5.4/0, 6.25/1, 7.1/2} {
    \node[patch] (b1_\ri_\ci) at (\c, \r) {};
  }
}
\node[font=\small, anchor=east] at (4.85, -3.95) {Frame $t{+}1$};
\foreach \r/\ri in {-3.15/0, -4.0/1, -4.85/2} {
  \foreach \c/\ci in {5.4/0, 6.25/1, 7.1/2} {
    \node[patch] (b2_\ri_\ci) at (\c, \r) {};
  }
}
\foreach \ci in {0, 1, 2} {
  \draw[->, red!70, line width=1.8pt] (b1_2_\ci.south) -- (b2_0_\ci.north);
}
\foreach \ri in {0, 1, 2} {
  \draw[->, blue!60, line width=1.4pt] (b1_\ri_0.east) -- (b1_\ri_1.west);
  \draw[->, blue!60, line width=1.4pt] (b1_\ri_1.east) -- (b1_\ri_2.west);
  \draw[->, blue!60, line width=1.4pt] (b2_\ri_0.east) -- (b2_\ri_1.west);
  \draw[->, blue!60, line width=1.4pt] (b2_\ri_1.east) -- (b2_\ri_2.west);
}

\node[font=\normalsize\bfseries] at (12.0, 1.6) {(c) Raster};
\node[font=\small, anchor=east] at (10.25, -0.8) {Frame $t$};
\foreach \r/\ri in {0/0, -0.85/1, -1.7/2} {
  \foreach \c/\ci in {10.8/0, 11.65/1, 12.5/2} {
    \node[patch] (c1_\ri_\ci) at (\c, \r) {};
  }
}
\node[font=\small, anchor=east] at (10.25, -3.95) {Frame $t{+}1$};
\foreach \r/\ri in {-3.15/0, -4.0/1, -4.85/2} {
  \foreach \c/\ci in {10.8/0, 11.65/1, 12.5/2} {
    \node[patch] (c2_\ri_\ci) at (\c, \r) {};
  }
}
\draw[->, blue!60, line width=1.4pt] (c1_0_0) -- (c1_0_1);
\draw[->, blue!60, line width=1.4pt] (c1_0_1) -- (c1_0_2);
\draw[->, blue!60, line width=1.4pt] (c1_0_2.south) to[out=-100,in=10] (c1_1_0.east);
\draw[->, blue!60, line width=1.4pt] (c1_1_0) -- (c1_1_1);
\draw[->, blue!60, line width=1.4pt] (c1_1_1) -- (c1_1_2);
\draw[->, blue!60, line width=1.4pt] (c1_1_2.south) to[out=-100,in=10] (c1_2_0.east);
\draw[->, blue!60, line width=1.4pt] (c1_2_0) -- (c1_2_1);
\draw[->, blue!60, line width=1.4pt] (c1_2_1) -- (c1_2_2);
\draw[->, orange!80!red, line width=2pt, dashed] (c1_2_2.south) to[out=-70,in=70] (c2_0_0.north);
\draw[->, blue!60, line width=1.4pt] (c2_0_0) -- (c2_0_1);
\draw[->, blue!60, line width=1.4pt] (c2_0_1) -- (c2_0_2);
\draw[->, blue!60, line width=1.4pt] (c2_0_2.south) to[out=-100,in=10] (c2_1_0.east);
\draw[->, blue!60, line width=1.4pt] (c2_1_0) -- (c2_1_1);
\draw[->, blue!60, line width=1.4pt] (c2_1_1) -- (c2_1_2);
\draw[->, blue!60, line width=1.4pt] (c2_1_2.south) to[out=-100,in=10] (c2_2_0.east);
\draw[->, blue!60, line width=1.4pt] (c2_2_0) -- (c2_2_1);
\draw[->, blue!60, line width=1.4pt] (c2_2_1) -- (c2_2_2);

\draw[->, red!70, line width=1.8pt] (2.5, -6.0) -- (3.4, -6.0) node[right, font=\small] {temporal pair ($\mathcal{L}_{\text{temp}}$)};
\draw[->, blue!60, line width=1.4pt] (6.5, -6.0) -- (7.4, -6.0) node[right, font=\small] {spatial pair ($\mathcal{L}_{\text{spat}}$)};
\draw[->, orange!80!red, line width=2pt, dashed] (10.5, -6.0) -- (11.4, -6.0) node[right, font=\small] {frame boundary};
\end{tikzpicture}
\caption{Contrastive pairing structures for the three spatio-temporal SAE variants. \textbf{(a) Temporal}: pairs the same spatial position across consecutive frames. \textbf{(b) Separate}: applies independent temporal (red, cross-frame) and spatial (blue, within-frame) contrastive losses. \textbf{(c) Raster}: serializes patches row-by-row, then across frames; at each frame boundary (orange dashed) the loss couples the last patch of frame $t$ to the first of frame $t{+}1$ --- a cross-frame boundary pair, not a same-position temporal correspondence.}
\vspace{-2mm}
\label{fig:variants}
\end{figure*}

Figure~\ref{fig:variants} illustrates the contrastive pairing structure for all three variants.

\paragraph{Matryoshka Hierarchical Grouping (+M)}
Orthogonal to the contrastive losses, we apply Matryoshka BatchTopK~\cite{bussmann2024batchtopksparseautoencoders,bussmann2025matryoshka} to partition the $H{=}6{,}144$-dimensional dictionary into a high-level group (features $0{-}1{,}227$, 20\% of the dictionary) and a low-level group (features $1{,}228{-}6{,}143$, 80\%). Both groups are trained jointly via BatchTopK with shared $k{=}64$. The high-level group is encouraged to be self-sufficient via an auxiliary reconstruction loss: $\mathcal{L}_{\text{mat}} = \alpha_{\text{mat}} \| \mathbf{x} - \hat{\mathbf{x}}_{\text{high}} \|_2^2$ where $\hat{\mathbf{x}}_{\text{high}}$ is decoded using only high-level features and $\alpha_{\text{mat}} = 0.1$. The motivation is to concentrate action-relevant, semantic features into a compact subset while relegating fine-grained details to the remaining features. We ablate the split ratio (5/95, 20/80, 50/50, 80/20) in the Supplementary Material. This can be combined with any contrastive variant (denoted +M, e.g., Temporal+M, Separate+M).

Both the contrastive and Matryoshka objectives compete with reconstruction for the SAE's limited sparse capacity, so the contrastive weight ($\lambda$ or temperature $\tau$) acts as a dial between reconstruction fidelity and \textit{task-oriented} encoding. We ablate this trade-off in Section~\ref{sec:ablation}.

\section{Experimental Setup and Baseline Analysis}
\label{sec:setup}

We extract features from \textbf{DINOv2 ViT-B/14}~\cite{oquab2024dinov}, a self-supervised image model giving $(16, 256, 768)$ features per clip, and \textbf{VideoMAE-B}~\cite{tong2022videomaemaskedautoencodersdataefficient}, a self-supervised video model pretrained on SSv2 giving $(8, 196, 768)$ --- a pairing that contrasts image-pretrained against video-pretrained representations. Main results use the final layer (layer 11); Section~\ref{sec:layers} additionally analyzes layers 3 and 7, which together sample the early, middle, and final stages of the 12-layer ViT (low-level appearance, mid-level part and motion structure, high-level semantics), the standard choice for layer-wise probing.

\paragraph{Datasets.}
(1) \textbf{Something-Something v2 (SSv2)}~\cite{goyal2017somethingsomethingvideodatabase}: 220k clips of fine-grained hand-object interactions with 174 action templates (e.g., ``Pushing [something] from left to right''). The action labels provide semantic supervision for evaluating feature quality. (2) \textbf{Kinetics-400 (K400)}~\cite{kay2017kineticshumanactionvideo}: 400 action classes with diverse scenes and appearance variation, testing generalization beyond fine-grained actions.

All SAEs use expansion factor 8$\times$ ($H = 6{,}144$), TopK with $k=64$, auxiliary loss coefficient $\alpha = 0.03$, and 10 epochs on 10{,}000 clips. Standard SAEs use batch size 4{,}096 on flattened tokens; clip-based variants use batch size 8 on whole clips. We set $\lambda_t = 0.1$, $\lambda_s = 0.05$, $\tau = 0.1$. All experiments use 3 random seeds with cross-seed variance below 1\%, so we report means. On A100 GPUs, training takes $\sim$30 min (standard), $\sim$60 min (clip-based contrastive, from per-clip loss computation), and $\sim$25 min (raster, pair-based).

\paragraph{Evaluation Metrics.}
We evaluate along seven axes:
\textbf{(1) Reconstruction quality}: $R^2$ (variance explained).
\textbf{(2) Temporal coherence}: mean lag-1 autocorrelation of SAE features across frames.
\textbf{(3) MonoSemanticity Score (MS)}: activation-weighted pairwise cosine similarity of each feature's top-activating clips, computed in the corresponding backbone's raw feature space. We adopt this metric, together with the surrounding evaluation protocol, directly from Pach et al.~\cite{pach2025sparse}, who introduced both for image SAEs; our contribution here is not the metric but the cross-backbone control in Section~\ref{sec:interpretability} that exposes its alignment artifact.
\textbf{(4) Action purity}: fraction of a feature's top-$k$ activating clips sharing the dominant action template.
\textbf{(5) Linear probe}: top-1 accuracy of logistic regression on SAE features for action classification.
\textbf{(6) Sparsity}: $L_0$ and dead feature fraction.
\textbf{(7) Feature uniqueness}: Jaccard index between top-activating clip sets across features (lower = more unique).

\subsection{Standard SAE Analysis}
\label{sec:interpretability}

Both backbones produce well-behaved SAEs on SSv2 (10K clips, $8\times$ expansion). SAE features outperform raw features on action classification for both models, with exact sparsity ($L_0 = 64$) and no dead features. Jaccard indices below 0.01 confirm that features partition the input space into distinct, non-overlapping concepts.

\paragraph{Cross-backbone monosemanticity validation.}
Under own-space evaluation, VideoMAE features achieve MS
$\approx 0.97$, compared to DINOv2's $0.32$--$0.37$, a gap of
roughly $2.7\times$. We investigate whether this reflects genuine interpretability differences or metric alignment. The MS metric computes clip similarity in each model's own raw feature space, following the evaluation structure of Pach et al.~\cite{pach2025sparse}. Because DINOv2 and VideoMAE induce different similarity geometries, their absolute own-space MS values need not be directly comparable. We recompute MS using CLIP ViT-B/32 as an independent similarity backbone (Table~\ref{tab:ms_cross}).

\begin{table}[t]
\centering
\caption{MonoSemanticity Score under different similarity backbones. Across all variants, VideoMAE reaches own-MS $0.969$--$0.974$ against DINOv2's $0.317$--$0.366$, a gap of roughly $2.7\times$; under an independent CLIP-based similarity space every entry collapses to $0.67$--$0.68$, revealing a metric alignment artifact. Own-MS is comparable \emph{within} a backbone but not \emph{across} backbones.}
\vspace{-2mm}
\label{tab:ms_cross}
\small
\begin{tabular}{llcc}
\toprule
Backbone & Variant & MS (own) & MS (CLIP) \\
\midrule
DINOv2 & Standard & 0.365 & 0.681 \\
DINOv2 & Temporal & 0.327 & 0.671 \\
DINOv2 & Separate & 0.328 & 0.671 \\
DINOv2 & Raster & 0.366 & 0.683 \\
DINOv2 & Temporal+M & 0.326 & 0.670 \\
DINOv2 & Separate+M & 0.317 & 0.668 \\
\midrule
VideoMAE & Standard & 0.973 & 0.680 \\
VideoMAE & Temporal & 0.973 & 0.678 \\
VideoMAE & Separate & 0.973 & 0.679 \\
VideoMAE & Raster & 0.974 & 0.681 \\
VideoMAE & Temporal+M & 0.971 & 0.677 \\
VideoMAE & Separate+M & 0.969 & 0.675 \\
\bottomrule
\end{tabular}
\end{table}

Under CLIP similarity, both backbones achieve MS $\approx$ 0.68 across all variants, and the gap closes. We do not claim CLIP is an objectively neutral judge; the point is that the gap is not stable under a similarity space chosen independently of either backbone, which suggests it is substantially influenced by differences between the backbones' native similarity geometries rather than by monosemanticity alone. This is a limitation of the Pach et al.\ protocol: it supports comparison \emph{within} a backbone, where any such bias is constant, but not \emph{across} backbones. We retain DINOv2-based MS for within-backbone comparisons in Table~\ref{tab:main_results}.

\begin{figure*}[t]
\centering
\includegraphics[width=0.85\textwidth]{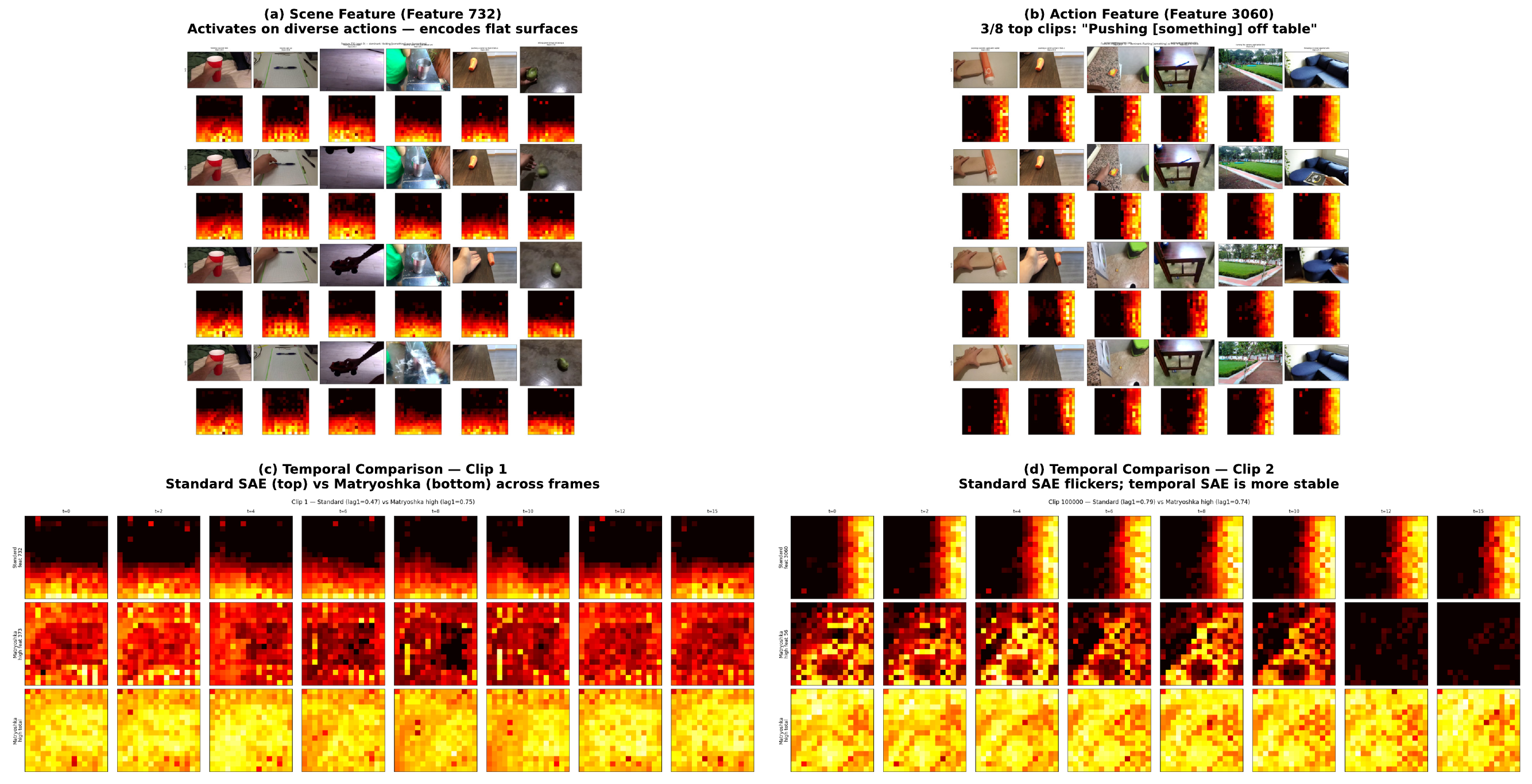}
\caption{SAE features on DINOv2/SSv2. \textbf{(a)} A scene feature activating on flat surfaces across diverse actions (holding, pushing, pouring, dropping). \textbf{(b)} An action-correlated feature firing on ``Pushing [something] off the table,'' with spatial activation concentrated on the pushing hand. \textbf{(c, d)} Temporal comparison: standard SAE features flicker across frames, while spatio-temporal SAE variants maintain more consistent activations.}
\vspace{-2mm}
\label{fig:features}
\end{figure*}

Figure~\ref{fig:features} illustrates the three feature categories and the temporal coherence problem: standard SAE features flicker across frames while Matryoshka features maintain stable activations.

\subsection{The Temporal Coherence Problem}
\label{sec:temporal_problem}

While standard SAEs produce interpretable features, they suffer from a fundamental limitation when applied to video: they destroy temporal coherence. Because each patch token is encoded independently, the top-$k$ selection mechanism can choose entirely different active features for the same spatial location across consecutive frames, even when the underlying visual content changes only slightly.

We measure temporal coherence as the mean lag-1 autocorrelation of SAE features across consecutive frames (Table~\ref{tab:main_results}). Raw DINOv2 features have lag-1 $= 0.435$; after SAE encoding, this drops to $0.277$ (a 36\% reduction). Temporal contrastive training partially recovers coherence (0.299). In practice, this means a feature representing ``hand pushing object'' may fire in frame~3, disappear in frame~4, and reappear in frame~6, making concept tracking across time unreliable. The severity varies by layer (lag-1 up to 0.75 at layer 3; see Section~\ref{sec:layers}).

\paragraph{Do simple fixes suffice?}
\label{sec:simple_baselines}

Before evaluating learned spatio-temporal objectives, we ask: can simple post-hoc smoothing resolve the temporal incoherence without modifying SAE training? We consider two baselines: \textbf{EMA smoothing} ($\tilde{\mathbf{z}}_{t} = \alpha \mathbf{z}_{t} + (1-\alpha) \tilde{\mathbf{z}}_{t-1}$) and \textbf{Temporal Union TopK} (reselecting top-$k$ from the union of active features at frames $t$ and $t{-}1$). Both appear in Table~\ref{tab:main_results} alongside the learned objectives.

Post-hoc baselines improve coherence by construction (EMA reaches lag-1 $= 0.28$) but degrade discrimination severely (probe drops from 19.1\% to 7.1\%). In contrast, temporal contrastive training improves \textit{both} simultaneously (lag-1 $0.299$, probe $20.2\%$). The benefit of learned objectives is not just temporal smoothing, which post-hoc methods achieve, but learning a dictionary that is \textit{jointly} coherent and discriminative.

\section{Spatio-Temporal SAE Variants}
\label{sec:st_results}

We now evaluate the spatio-temporal SAE variants from Section~\ref{sec:method_st}. We compare against raw backbone features, a standard TopK SAE, post-hoc temporal smoothing, a dimension-matched reconstructed SAE, and random-feature ablation. These controls isolate the effects of sparse decomposition, temporal training, representation dimensionality, and learned dictionary structure, respectively.

\subsection{Full Variant Comparison}
\label{sec:ablation}

We sweep the contrastive weight ($\lambda \in \{0.01, 0.05, 0.5\}$, $\tau \in \{0.05, 0.2, \\ 0.5\}$) across all five variants on DINOv2/SSv2 (29 configurations; full ablation in the Supplementary Material). Table~\ref{tab:main_results} shows the best configuration per variant at 10K. All variants have zero dead features.

\begin{table}[t]
\centering
\caption{Balanced SAE variants on DINOv2/SSv2 (10K clips), selected from 29 configurations over $\lambda \in \{0.01,0.05,0.5\}$ and $\tau \in \{0.05,0.2,0.5\}$; full results are in the Supplementary Material. Higher contrastive weight can further improve probe accuracy at the cost of $R^2$; Separate at $\lambda{=}0.5$ reaches $21.3\%$ (Table~\ref{tab:causal_ablation}). All variants outperform raw features with zero dead features. Probe uses an 80/20 split. $^\dagger$Coherence-optimized: lag-1 exceeds raw DINOv2.}
\vspace{-2mm}
\label{tab:main_results}
\begin{tabular}{lcccc}
\toprule
Variant & $R^2 \uparrow$ & Lag-1 $\uparrow$ & MS $\uparrow$ & Probe $\uparrow$ \\
\midrule
Raw DINOv2 & --- & 0.435 & --- & 17.4\% \\
Standard SAE & \textbf{0.454} & 0.277 & 0.359 & 19.1\% \\
\midrule
\multicolumn{5}{l}{\textit{Post-hoc smoothing of the standard SAE}} \\
\quad EMA ($\alpha{=}0.3$) & --- & 0.280 & --- & 7.1\% \\
\quad EMA ($\alpha{=}0.5$) & --- & 0.210 & --- & 7.8\% \\
\quad Temporal Union & --- & 0.150 & --- & 8.2\% \\
\midrule
\multicolumn{5}{l}{\textit{Learned spatio-temporal objectives (ours)}} \\
Temporal & 0.320 & 0.299 & 0.337 & 20.2\% \\
Separate & 0.318 & 0.290 & 0.333 & 20.2\% \\
Raster & 0.418 & 0.306 & \textbf{0.375} & 18.3\% \\
Temporal+M & 0.334 & 0.360 & 0.334 & \textbf{20.4\%} \\
Separate+M & 0.310 & 0.358 & 0.326 & 19.7\% \\
Temporal+M$^\dagger$ & 0.281 & \textbf{0.462} & 0.344 & 20.0\% \\
\bottomrule
\end{tabular}
\end{table}

\paragraph{All SAE variants outperform raw features.}
The standard SAE (19.1\%) surpasses raw DINOv2 (17.4\%), and contrastive variants reach 20.4\% at the balanced operating points of Table~\ref{tab:main_results}. Trading further reconstruction for discrimination pushes this higher still: Separate at $\lambda{=}0.5$ attains $21.3\%$, $+3.9$ pp over raw features and the best probe accuracy we observe at 10K. These are frozen-feature linear probes, used to measure how much action-relevant structure survives sparse decomposition rather than to claim state-of-the-art action recognition. The same trend holds at 20K clips, where probe accuracy rises for every variant (standard $19.1\% \to 21.9\%$, temporal $20.2\% \to 22.8\%$); see the Supplementary Material.

\paragraph{The reconstruction--coherence trade-off is controllable.}
All contrastive variants sacrifice reconstruction relative to the standard SAE ($R^2 = 0.454$). At low contrastive weight ($\lambda{=}0.01$), all variants maintain positive $R^2$ ($0.31{-}0.42$). At high weight, $R^2$ decreases while temporal coherence increases. The full 29-configuration ablation (Supplementary Material) shows this is a smooth, monotonic trade-off controlled by a single hyperparameter.

\paragraph{Matryoshka variants are the strongest overall.}
Temporal+M achieves the best probe accuracy at the balanced settings (20.4\%), the highest temporal coherence among contrastive variants (lag-1 $= 0.360$ at $\lambda{=}0.01$), and at $\tau{=}0.2$ \textit{exceeds} raw DINOv2 coherence (lag-1 $= 0.462$ vs.\ 0.435) --- the only variant to clear the raw backbone on this metric. The hierarchical structure concentrates action-relevant features in the high-level group while delegating fine-grained texture to the low-level group. This separation allows the contrastive loss to improve temporal coherence without sacrificing discrimination. We recommend Temporal+M as the default variant for applications that require both interpretability and action-discriminative features.

\paragraph{Temporal and Separate offer balanced discrimination.}
Temporal and Separate reach the same probe accuracy at their balanced settings (20.2\%), but Temporal concentrates predictive signal into the fewest features: just 10 ablated features cause a $-9.3$ pp drop in the targeted ablation (Table~\ref{tab:causal_ablation}). Separate adds independent spatial consistency, distributes predictive signal more broadly, and is the variant that benefits most from a large contrastive weight ($21.3\%$ at $\lambda{=}0.5$). Both are well-suited when Matryoshka's hierarchical structure is unnecessary and a single contrastive loss suffices.

\paragraph{Raster excels at reconstruction and monosemanticity.}
Among the contrastive variants Raster preserves the highest $R^2$ (0.418 at $\lambda{=}0.01$, closest to the standard SAE's 0.454) and achieves the best MS score of any configuration we train (0.433 at $\lambda{=}0.5$). Its pair-based loss applies only \textit{local} contrastive pressure between adjacent patches, disrupting reconstruction less than the clip-level objectives, which enforce global consistency across all frames. That same local consistency acts as an inductive bias for monosemanticity: each feature must agree across spatially adjacent and cross-frame raster pairs. It also yields the purest features by action template: across the top-50 features, mean action purity rises from 1.36/8 (standard) to 1.60/8 (temporal) to 1.74/8 (raster), with raster reaching 4/8 against the standard SAE's 3/8. Probe accuracy nonetheless drops at high $\lambda$, as local consistency redirects capacity away from global action discrimination. Raster therefore suits applications prioritizing reconstruction fidelity or interpretability over classification.

\subsection{Targeted Feature Ablation}

To test whether SAE features are \textit{functionally important} to the probe's action predictions (not merely correlated with them), we perform a targeted feature ablation following Stevens et al.~\cite{stevens2025interpretabletestablevisionfeatures}. We train a linear probe once on full SAE features, identify the top-$N$ most important features by probe weight magnitude, zero them out, and evaluate with the \textit{same probe weights} (no retraining). As a control, we ablate $N$ random features.

\begin{table*}[t]
\centering
\caption{Targeted feature ablation on DINOv2/SSv2 (10K, no probe retraining, following Stevens et al.~\cite{stevens2025interpretabletestablevisionfeatures}). We train a probe once, then zero the top-$N$ features by weight magnitude and evaluate with the same weights. Baseline accuracies ($N{=}0$) differ from Table~\ref{tab:main_results} because each variant uses the checkpoint with the strongest ablation sensitivity (Separate: $\lambda{=}0.5$, Raster: $\tau{=}0.5$, Temp.+M / Sep.+M: $\lambda{=}0.01$), not the best-probe configuration.}
\vspace{-2mm}
\label{tab:causal_ablation}
\begin{tabular}{lcccccccccccc}
\toprule
\multirow{2}{*}{$N$} & \multicolumn{2}{c}{Standard} & \multicolumn{2}{c}{Temporal} & \multicolumn{2}{c}{Separate} & \multicolumn{2}{c}{Raster} & \multicolumn{2}{c}{Temp.+M} & \multicolumn{2}{c}{Sep.+M} \\
\cmidrule(lr){2-3}\cmidrule(lr){4-5}\cmidrule(lr){6-7}\cmidrule(lr){8-9}\cmidrule(lr){10-11}\cmidrule(lr){12-13}
& Top & Rand & Top & Rand & Top & Rand & Top & Rand & Top & Rand & Top & Rand \\
\midrule
0 & \multicolumn{2}{c}{19.1} & \multicolumn{2}{c}{20.2} & \multicolumn{2}{c}{21.3} & \multicolumn{2}{c}{19.3} & \multicolumn{2}{c}{18.8} & \multicolumn{2}{c}{19.7} \\
10 & 18.1 & 19.2 & \textbf{10.9} & 20.4 & 19.6 & 21.1 & 17.8 & 19.3 & 16.4 & 18.9 & \textbf{12.9} & 19.7 \\
50 & 16.3 & 19.2 & 13.4 & 20.1 & 16.9 & 21.1 & \textbf{13.3} & 19.3 & 13.0 & 18.8 & \textbf{11.0} & 19.5 \\
100 & 17.4 & 19.2 & 16.1 & 19.9 & 15.9 & 20.9 & \textbf{12.8} & 18.5 & 12.9 & 18.6 & 12.4 & 19.1 \\
500 & 17.3 & 18.7 & 14.1 & 17.4 & 14.4 & 19.4 & \textbf{8.5} & 14.9 & 10.9 & 17.1 & \textbf{10.3} & 16.4 \\
\bottomrule
\end{tabular}
\end{table*}

Table~\ref{tab:causal_ablation} reveals clear differences in how variants distribute predictive information. The \textbf{standard SAE} shows moderate redundancy: ablating the top 50 features drops accuracy by 2.8 pp (19.1\% to 16.3\%), while ablating 50 random features leaves accuracy essentially unchanged (19.2\%). The non-monotonicity at larger $N$ (e.g., partial recovery at $N{=}100$) reflects interactions between positively and negatively weighted features in the multiclass probe. The \textbf{temporal contrastive SAE} shows extreme concentration at the top: removing just 10 features (0.16\% of the dictionary) drops accuracy from 20.2\% to 10.9\% ($-9.3$ pp). \textbf{Raster} shows the strongest sustained drop: top-500 ablation reduces accuracy from 19.3\% to 8.5\% ($-10.8$ pp), more than halving it. \textbf{Separate+M} is similarly concentrated ($-9.4$ pp at $N{=}500$). These results show that contrastive and Matryoshka training concentrate the probe's predictive weight onto specific identifiable features rather than distributing it broadly. The concentration pattern varies by variant: temporal focuses signal into a handful of features, while raster and Separate+M spread it across a larger but still identifiable subset.

The mechanism is the pairing structure. The temporal loss makes the same features fire across consecutive frames at one spatial position, creating a small set of ``anchor'' features that the probe weights heavily --- efficient, but fragile to targeted ablation. Raster's unified loss instead enforces consistency across all adjacent pairs, including within-frame neighbors, spreading importance over a larger set of moderately weighted features. Separate+M sits between the two: Matryoshka concentrates action-relevant features into the high-level group, but importance is distributed across that whole group rather than a handful of individual features.

\subsection{Cross-Dataset and Cross-Model Validation}
\label{sec:crossdataset}

Full cross-model validation (DINOv2 vs.\ VideoMAE), the Matryoshka split analysis, and the complete cross-dataset tables are provided in the Supplementary Material; we summarize the outcome here.

Spatio-temporal objectives are most beneficial for DINOv2, which lacks temporal pretraining, whereas the contrastive term leaves VideoMAE's probe accuracy essentially unchanged (we return to this rule in Section~\ref{sec:discussion}). The Matryoshka 20/80 split concentrates action-relevant information into 20\% of the dictionary (19.6\% probe using only 1,228 of 6,144 features). The temporal objective also partially transfers to K400. On SSv2 the SAE improves over raw features for both backbones ($17.4 \rightarrow 20.2$ for DINOv2, $29.2 \rightarrow 32.4$ for VideoMAE). On K400 the SAE trails raw features ($33.3 \rightarrow 28.3$ for DINOv2), but temporal contrastive training recovers much of the deficit ($33.3 \rightarrow 30.5$), reducing the gap from $-5.0$ to $-2.8$ pp, a 44\% reduction. K400's larger camera motion, which weakens the fixed-position pairing assumption, is the most likely cause of the residual gap.

\subsection{Text-Video Retrieval with SAE Features}
\label{sec:retrieval}

To evaluate SAE features in a multimodal setting, we conduct a text-video retrieval experiment. We use SSv2 action templates as text queries, encode them with CLIP ViT-B/32~\cite{radford2021learningtransferablevisualmodels}, and learn a Ridge-regularized linear projection from video to text space with an 80/20 train/test split (Ridge $\alpha$ selected via 5-fold cross-validation on the training split; $\alpha=1.0$ selected). We report the best-performing contrastive variant per backbone (Temporal for DINOv2, Raster for VideoMAE) alongside Standard and raw baselines. Table~\ref{tab:retrieval} includes a dimensionality control: SAE-reconstructed features (768-dim, same as raw) isolate the effect of sparse decomposition from the overcomplete representation.

\begin{table}[t]
    \centering
    \caption{Text-video retrieval on SSv2 (10K clips, 80/20 split, Ridge $\alpha{=}1.0$). SAE-reconstructed (768d) controls for dimensionality. VideoMAE SAE features achieve the highest absolute R@1.}
\vspace{-2mm}
    \label{tab:retrieval}
    \begin{tabular}{llcc}
    \toprule
    Model & Representation & R@1 & R@5 \\
    \midrule
    \multirow{4}{*}{DINOv2} & Raw (768$\to$512) & 4.6\% & 15.2\% \\
    & Reconstructed (768$\to$512) & 5.4\% & 16.6\% \\
    & Standard SAE (6144$\to$512) & 13.0\% & 27.8\% \\
    & Temporal Contr. (6144$\to$512) & \textbf{13.1}\% & \textbf{28.3}\% \\
    \midrule
    \multirow{4}{*}{VideoMAE} & Raw (768$\to$512) & 13.4\% & 26.6\% \\
    & Reconstructed (768$\to$512) & 15.2\% & 27.6\% \\
    & Standard SAE (6144$\to$512) & \textbf{25.6\%} & \textbf{43.0\%} \\
    & Raster Contr. (6144$\to$512) & 25.1\% & 42.2\% \\
    \bottomrule
    \end{tabular}
\end{table}

\begin{figure*}[t]
\centering
\includegraphics[width=0.85\textwidth]{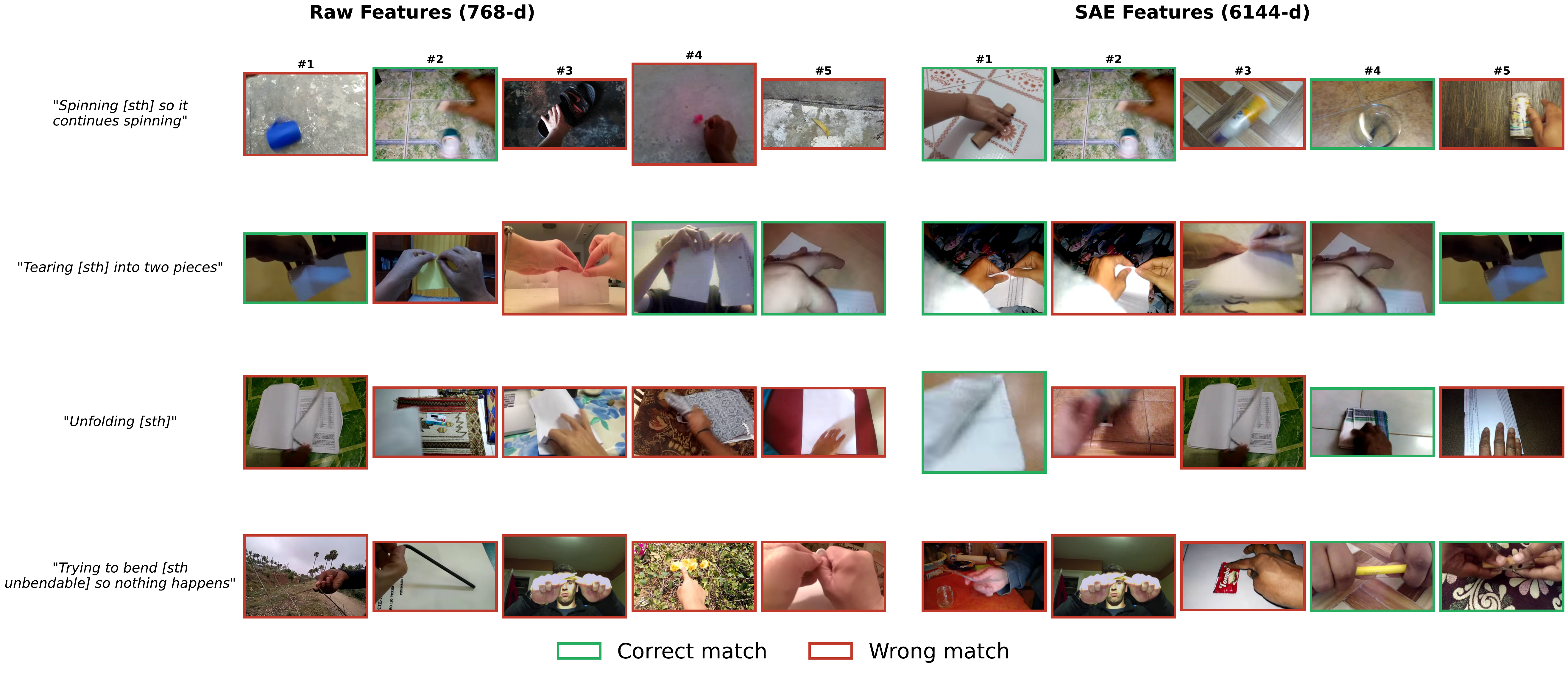}
\caption{Qualitative text-video retrieval comparison (DINOv2/SSv2, Standard SAE). For each query, we show the top-5 retrieved clips using raw features (left) vs.\ SAE features (right). Green borders indicate exact template match (conservative: visually similar clips with different templates count as incorrect). SAE features retrieve 11/20 correct matches vs.\ 4/20 for raw features.}
\vspace{-2mm}
\label{fig:retrieval}
\end{figure*}

SAE features dramatically improve retrieval on both backbones. For DINOv2, standard SAE achieves 13.0\% R@1, which is 2.8$\times$ higher than raw features (4.6\%). For VideoMAE, the improvement is even larger: 25.6\% R@1 vs.\ 13.4\% raw (1.9$\times$), achieving the highest absolute retrieval accuracy. Critically, the dimensionality control confirms the effect is not merely from the overcomplete representation: at the \textit{same 768 dimensions}, SAE-reconstructed features outperform raw on both DINOv2 (5.4\% vs.\ 4.6\%) and VideoMAE (15.2\% vs.\ 13.4\%). This indicates that sparse decomposition genuinely improves the alignment between visual and textual action semantics, likely because each SAE feature encodes a single concept that maps cleanly to words. Figure~\ref{fig:retrieval} illustrates this qualitatively: for four action template queries, SAE features retrieve visually relevant clips while raw features largely fail.

\subsection{Layer Analysis}
\label{sec:layers}

We analyze layers 3, 7, and 11 for both backbones (full tables in the Supplementary Material). For DINOv2, action discrimination increases with depth, and temporal contrastive training is most beneficial at the final layer. At layer 7 it instead degrades probe accuracy (11.3\% $\to$ 2.4\%). VideoMAE layer 7 reaches 32.9\%, while temporal contrastive training has little effect across layers, likely because temporal pretraining already provides coherent features. Layer-wise feature sets were extracted separately, so we do not overinterpret absolute cross-layer differences.

\section{Discussion}
\label{sec:discussion}

\paragraph{Hard TopK vs.\ soft sparsity}
Our results trace the temporal coherence problem to the hard top-$k$ selection. Small input perturbations between consecutive frames can flip which $k$ features are active. To test whether \textit{soft} sparsity mechanisms can address this directly, we train SAE variants using sparsemax~\cite{pmlr-v48-martins16} and entmax-1.5~\cite{peters-etal-2019-sparse} activations, sweeping the sparsemax temperature from $0.1$ to $50$ (full table in the Supplementary Material). Raising the temperature raises $L_0$ from $1.0$ to $20.2$ and recovers probe accuracy from $2.0\%$ to $19.8\%$, surpassing raw features. But even at $T{=}50$, $L_0$ remains far below TopK's $64$ and $R^2$ stays negative ($-0.32$) against the matched hard-TopK control for this ablation ($R^2 = 0.309$; that control is TopK with the same temporal contrastive term, not the standard SAE of Table~\ref{tab:main_results}). Temperature scaling alone therefore cannot bridge soft and hard sparsity; reaching TopK-level $L_0$ with smooth activations likely requires hybrid approaches such as soft selection with a hard floor. We keep TopK and add a training-time term that makes its selections temporally consistent, which costs nothing in interpretability: it pushes the \textit{same} features to fire across adjacent frames. The choice of $k{=}64$ follows Gao et al.~\cite{gao2024scalingevaluatingsparseautoencoders}.

\paragraph{Learned vs.\ random dictionaries.}
To verify that SAE dictionaries learn meaningful structure rather than acting as random overcomplete projections, we train a control SAE with frozen random decoder weights (same architecture, only the encoder is optimized). At 10K clips, the learned decoder achieves $R^2 = 0.910$ (clip-pooled) compared to $R^2 = 0.578$ for the frozen decoder. This large gap confirms that the decoder learns a structured dictionary. Probe accuracy is similar (19.7\% learned vs.\ 19.6\% frozen), indicating that the encoder extracts discriminative features regardless of decoder quality. The decoder's role is reconstruction fidelity, not discrimination.

\paragraph{Practical recommendations.}
Table~\ref{tab:goals} gives four operating points, one per goal. The choice is also \textit{backbone-conditional}. Contrastive objectives are most useful on image-pretrained backbones such as DINOv2, which have no temporal pretraining to inherit structure from, whereas temporally pretrained backbones such as VideoMAE benefit mainly from the sparse decomposition itself.

\begin{table}[t]
\centering
\caption{Which variant to use for which goal (DINOv2/SSv2, 10K clips). No variant dominates every metric; the contrastive weight selects the operating point.}
\vspace{-2mm}
\label{tab:goals}
\small
\begin{tabular}{llc}
\toprule
Goal & Variant & Achieved \\
\midrule
Reconstruction fidelity & Raster, $\lambda{=}0.01$ & $R^2 = 0.418$ \\
Temporal coherence & Temporal+M, $\tau{=}0.2$ & lag-1 $= 0.462$ \\
Action discrimination & Separate, $\lambda{=}0.5$ & probe $= 21.3\%$ \\
Interpretability (MS) & Raster, $\lambda{=}0.5$ & MS $= 0.433$ \\
\bottomrule
\end{tabular}
\end{table}

\paragraph{Toward interpretable video-language assistants.}
The target for this work is the visual encoder inside a multimodal assistant, whose failures are often temporal: an object or on-screen element is represented correctly in one frame and lost in the next. The coherence loss documented above means the feature firing on an entity at frame $t$ need not be the one firing at $t{+}1$, which makes concept-level intervention ill-posed across a clip. Temporally coherent dictionaries restore that well-posedness, and combined with the known steerability of vision-encoder SAEs~\cite{pach2025sparse} they suggest stable handles on an assistant's visual stream. Extending this to the visual subspaces of omni-modal assistants is the natural next step.

\paragraph{Limitations.}
Our temporal losses pair the same spatial position across consecutive frames, assuming approximate spatial correspondence. SSv2's stationary cameras satisfy this for the vast majority of clips; K400's greater camera motion does not, which likely explains its smaller contrastive gains. When the assumption is violated the loss receives noisy positive pairs, weakening but not collapsing the signal, as InfoNCE tolerates moderate label noise. Optical-flow correspondences~\cite{teed2020raftrecurrentallpairsfield} could address this at the cost of external dependencies and the model-agnostic property we value. We fix $k{=}64$ and expansion factor $8\times$ throughout; varying these may shift the frontier. Our retrieval uses SSv2's formulaic action templates, so results on naturalistic benchmarks (MSR-VTT, DiDeMo) may differ. Across 3 seeds the maximum observed standard deviation is 0.3 pp for probe accuracy and 0.005 for $R^2$, so borderline comparisons should not be overinterpreted.

\section{Conclusion}

\label{sec:conclusion}

We presented the first systematic study of Sparse Autoencoders on video representations. Standard SAEs decompose video into monosemantic concepts, but they destroy the temporal structure that makes those concepts trackable. The monosemanticity metric inherited from image SAEs also hides a backbone-alignment artifact, which does not survive an independent similarity space. Contrastive pairing along the temporal and spatial axes does not remove the tension between reconstruction and coherence. It turns that tension into a controllable axis. The contrastive weight selects where a dictionary sits on the frontier, and no variant dominates, so we report four operating points. The hard TopK selection remains the open problem: soft alternatives recover coherence only by surrendering reconstruction. Hybrid soft-hard sparsity and multiresolution decompositions are, in our view, the most promising directions.

\bibliographystyle{ACM-Reference-Format}
\bibliography{references}

@inproceedings{bussmann2025matryoshka,
  title={Learning Multi-Level Features with Matryoshka Sparse Autoencoders},
  author={Bart Bussmann and Noa Nabeshima and Adam Karvonen and Neel Nanda},
  booktitle={Proceedings of the 42nd International Conference on Machine Learning},
  series={Proceedings of Machine Learning Research},
  volume={267},
  pages={6077--6101},
  year={2025},
  publisher={PMLR},
  url={https://arxiv.org/abs/2503.17547},
}

@article{bricken2023monosemanticity,
  title={Towards Monosemanticity: Decomposing Language Models With Dictionary Learning},
  author={Bricken, Trenton and Templeton, Adly and Batson, Joshua and Chen, Brian and Jermyn, Adam and Conerly, Tom and Turner, Nick and Anil, Cem and Denison, Carson and Askell, Amanda and others},
  journal={Transformer Circuits Thread},
  year={2023}
}

@article{templeton2024scaling,
  title={Scaling Monosemanticity: Extracting Interpretable Features from Claude 3 Sonnet},
  author={Templeton, Adly and Conerly, Tom and Marcus, Jonathan and Lindsey, Jack and Bricken, Trenton and Chen, Brian and Pearce, Adam and Citro, Craig and Ameisen, Emmanuel and Jones, Andy and others},
  journal={Transformer Circuits Thread},
  year={2024}
}

@misc{goyal2017somethingsomethingvideodatabase,
      title={The "something something" video database for learning and evaluating visual common sense}, 
      author={Raghav Goyal and Samira Ebrahimi Kahou and Vincent Michalski and Joanna Materzyńska and Susanne Westphal and Heuna Kim and Valentin Haenel and Ingo Fruend and Peter Yianilos and Moritz Mueller-Freitag and Florian Hoppe and Christian Thurau and Ingo Bax and Roland Memisevic},
      year={2017},
      eprint={1706.04261},
      archivePrefix={arXiv},
      primaryClass={cs.CV},
      url={https://arxiv.org/abs/1706.04261}, 
}

@misc{kay2017kineticshumanactionvideo,
      title={The Kinetics Human Action Video Dataset}, 
      author={Will Kay and Joao Carreira and Karen Simonyan and Brian Zhang and Chloe Hillier and Sudheendra Vijayanarasimhan and Fabio Viola and Tim Green and Trevor Back and Paul Natsev and Mustafa Suleyman and Andrew Zisserman},
      year={2017},
      eprint={1705.06950},
      archivePrefix={arXiv},
      primaryClass={cs.CV},
      url={https://arxiv.org/abs/1705.06950}, 
}

@misc{radford2021learningtransferablevisualmodels,
      title={Learning Transferable Visual Models From Natural Language Supervision}, 
      author={Alec Radford and Jong Wook Kim and Chris Hallacy and Aditya Ramesh and Gabriel Goh and Sandhini Agarwal and Girish Sastry and Amanda Askell and Pamela Mishkin and Jack Clark and Gretchen Krueger and Ilya Sutskever},
      year={2021},
      eprint={2103.00020},
      archivePrefix={arXiv},
      primaryClass={cs.CV},
      url={https://arxiv.org/abs/2103.00020}, 
}

@misc{wang2024staaspatiotemporalattentionattribution,
      title={STAA: Spatio-Temporal Attention Attribution for Real-Time Interpreting Transformer-based Video Models}, 
      author={Zerui Wang and Yan Liu},
      year={2024},
      eprint={2411.00630},
      archivePrefix={arXiv},
      primaryClass={cs.CV},
      url={https://arxiv.org/abs/2411.00630}, 
}

@inproceedings{zhang2021contrastivespatiotemporalpretextlearning,
  title={Contrastive Spatio-Temporal Pretext Learning for Self-supervised Video Representation},
  author={Yujia Zhang and Lai-Man Po and Xuyuan Xu and Mengyang Liu and Yexin Wang and Weifeng Ou and Yuzhi Zhao and Wing-Yin Yu},
  booktitle={Proceedings of the AAAI Conference on Artificial Intelligence},
  volume={36},
  pages={3380--3389},
  year={2022},
}

@misc{cunningham2023sparseautoencodershighlyinterpretable,
      title={Sparse Autoencoders Find Highly Interpretable Features in Language Models}, 
      author={Hoagy Cunningham and Aidan Ewart and Logan Riggs and Robert Huben and Lee Sharkey},
      year={2023},
      eprint={2309.08600},
      archivePrefix={arXiv},
      primaryClass={cs.LG},
      url={https://arxiv.org/abs/2309.08600}, 
}

@misc{gao2024scalingevaluatingsparseautoencoders,
      title={Scaling and evaluating sparse autoencoders}, 
      author={Leo Gao and Tom Dupré la Tour and Henk Tillman and Gabriel Goh and Rajan Troll and Alec Radford and Ilya Sutskever and Jan Leike and Jeffrey Wu},
      year={2024},
      eprint={2406.04093},
      archivePrefix={arXiv},
      primaryClass={cs.LG},
      url={https://arxiv.org/abs/2406.04093}, 
}

@misc{rajamanoharan2024jumpingaheadimprovingreconstruction,
      title={Jumping Ahead: Improving Reconstruction Fidelity with JumpReLU Sparse Autoencoders}, 
      author={Senthooran Rajamanoharan and Tom Lieberum and Nicolas Sonnerat and Arthur Conmy and Vikrant Varma and János Kramár and Neel Nanda},
      year={2024},
      eprint={2407.14435},
      archivePrefix={arXiv},
      primaryClass={cs.LG},
      url={https://arxiv.org/abs/2407.14435}, 
}

@misc{bussmann2024batchtopksparseautoencoders,
      title={BatchTopK Sparse Autoencoders}, 
      author={Bart Bussmann and Patrick Leask and Neel Nanda},
      year={2024},
      eprint={2412.06410},
      archivePrefix={arXiv},
      primaryClass={cs.LG},
      url={https://arxiv.org/abs/2412.06410}, 
}

@misc{stevens2025interpretabletestablevisionfeatures,
      title={Interpretable and Testable Vision Features via Sparse Autoencoders}, 
      author={Samuel Stevens and Wei-Lun Chao and Tanya Berger-Wolf and Yu Su},
      year={2025},
      eprint={2502.06755},
      archivePrefix={arXiv},
      primaryClass={cs.CV},
      url={https://arxiv.org/abs/2502.06755}, 
}

@misc{joseph2025steeringclipsvisiontransformer,
      title={Steering CLIP's vision transformer with sparse autoencoders}, 
      author={Sonia Joseph and Praneet Suresh and Ethan Goldfarb and Lorenz Hufe and Yossi Gandelsman and Robert Graham and Danilo Bzdok and Wojciech Samek and Blake Aaron Richards},
      year={2025},
      eprint={2504.08729},
      archivePrefix={arXiv},
      primaryClass={cs.CV},
      url={https://arxiv.org/abs/2504.08729}, 
}

@inproceedings{
    lim2025sparse,
    title={Sparse autoencoders reveal selective remapping of visual concepts during adaptation},
    author={Hyesu Lim and Jinho Choi and Jaegul Choo and Steffen Schneider},
    booktitle={The Thirteenth International Conference on Learning Representations},
    year={2025},
    url={https://openreview.net/forum?id=imT03YXlG2}
}

@inproceedings{
    bhalla2026temporal,
    title={Temporal Sparse Autoencoders: Leveraging the Sequential Nature of Language for Interpretability},
    author={Usha Bhalla and Alex Oesterling and Claudio Mayrink Verdun and Himabindu Lakkaraju and Flavio Calmon},
    booktitle={The Fourteenth International Conference on Learning Representations},
    year={2026},
    url={https://openreview.net/forum?id=bojVI4l9Kn}
}

@misc{haon2025mechanisticinterpretabilitysteeringvisionlanguageaction,
      title={Mechanistic interpretability for steering vision-language-action models}, 
      author={Bear Häon and Kaylene Stocking and Ian Chuang and Claire Tomlin},
      year={2025},
      eprint={2509.00328},
      archivePrefix={arXiv},
      primaryClass={cs.RO},
      url={https://arxiv.org/abs/2509.00328}, 
}

@inproceedings{
    khan2025controlling,
    title={Controlling Vision{\textendash}Language{\textendash}Action Policies through Sparse Latent Directions},
    author={Momin Ahmad Khan and Novak Boskov and Fatima M. Anwar and Manzoor A. Khan},
    booktitle={Mechanistic Interpretability Workshop at NeurIPS 2025},
    year={2025},
    url={https://openreview.net/forum?id=wtf3ww1EOL}
}

@article{
    oquab2024dinov,
    title={{DINO}v2: Learning Robust Visual Features without Supervision},
    author={Maxime Oquab and Timoth{\'e}e Darcet and Th{\'e}o Moutakanni and Huy V. Vo and Marc Szafraniec and Vasil Khalidov and Pierre Fernandez and Daniel HAZIZA and Francisco Massa and Alaaeldin El-Nouby and Mido Assran and Nicolas Ballas and Wojciech Galuba and Russell Howes and Po-Yao Huang and Shang-Wen Li and Ishan Misra and Michael Rabbat and Vasu Sharma and Gabriel Synnaeve and Hu Xu and Herve Jegou and Julien Mairal and Patrick Labatut and Armand Joulin and Piotr Bojanowski},
    journal={Transactions on Machine Learning Research},
    issn={2835-8856},
    year={2024},
    url={https://openreview.net/forum?id=a68SUt6zFt},
    note={Featured Certification}
}

@inproceedings{peters-etal-2019-sparse,
    title = "Sparse Sequence-to-Sequence Models",
    author = "Peters, Ben  and
      Niculae, Vlad  and
      Martins, Andr{\'e} F. T.",
    editor = "Korhonen, Anna  and
      Traum, David  and
      M{\`a}rquez, Llu{\'i}s",
    booktitle = "Proceedings of the 57th Annual Meeting of the Association for Computational Linguistics",
    month = jul,
    year = "2019",
    address = "Florence, Italy",
    publisher = "Association for Computational Linguistics",
    url = "https://aclanthology.org/P19-1146/",
    doi = "10.18653/v1/P19-1146",
    pages = "1504--1519"
}

@misc{teed2020raftrecurrentallpairsfield,
      title={RAFT: Recurrent All-Pairs Field Transforms for Optical Flow}, 
      author={Zachary Teed and Jia Deng},
      year={2020},
      eprint={2003.12039},
      archivePrefix={arXiv},
      primaryClass={cs.CV},
      url={https://arxiv.org/abs/2003.12039}, 
}

@InProceedings{pmlr-v48-martins16,
  title = 	 {From Softmax to Sparsemax: A Sparse Model of Attention and Multi-Label Classification},
  author = 	 {Martins, Andre and Astudillo, Ramon},
  booktitle = 	 {Proceedings of The 33rd International Conference on Machine Learning},
  pages = 	 {1614--1623},
  year = 	 {2016},
  editor = 	 {Balcan, Maria Florina and Weinberger, Kilian Q.},
  volume = 	 {48},
  series = 	 {Proceedings of Machine Learning Research},
  address = 	 {New York, New York, USA},
  month = 	 {20--22 Jun},
  publisher =    {PMLR},
  url = 	 {https://proceedings.mlr.press/v48/martins16.html}
}

@misc{tong2022videomaemaskedautoencodersdataefficient,
      title={VideoMAE: Masked Autoencoders are Data-Efficient Learners for Self-Supervised Video Pre-Training}, 
      author={Zhan Tong and Yibing Song and Jue Wang and Limin Wang},
      year={2022},
      eprint={2203.12602},
      archivePrefix={arXiv},
      primaryClass={cs.CV},
      url={https://arxiv.org/abs/2203.12602}, 
}

@inproceedings{
    pach2025sparse,
    title={Sparse Autoencoders Learn Monosemantic Features in Vision-Language Models},
    author={Mateusz Pach and Shyamgopal Karthik and Quentin Bouniot and Serge Belongie and Zeynep Akata},
    booktitle={The Thirty-ninth Annual Conference on Neural Information Processing Systems},
    year={2025},
    url={https://openreview.net/forum?id=DaNnkQJSQf}
}

@article{Dave_2022,
   title={TCLR: Temporal contrastive learning for video representation},
   volume={219},
   ISSN={1077-3142},
   url={http://dx.doi.org/10.1016/j.cviu.2022.103406},
   DOI={10.1016/j.cviu.2022.103406},
   journal={Computer Vision and Image Understanding},
   publisher={Elsevier BV},
   author={Dave, Ishan and Gupta, Rohit and Rizve, Mamshad Nayeem and Shah, Mubarak},
   year={2022},
   month=jun, pages={103406} 
   }

@misc{oord2019representationlearningcontrastivepredictive,
      title={Representation Learning with Contrastive Predictive Coding}, 
      author={Aaron van den Oord and Yazhe Li and Oriol Vinyals},
      year={2019},
      eprint={1807.03748},
      archivePrefix={arXiv},
      primaryClass={cs.LG},
      url={https://arxiv.org/abs/1807.03748}, 
}

@inproceedings{manasyan2025temporallyconsistentobjectcentriclearning,
  title={Temporally Consistent Object-Centric Learning by Contrasting Slots},
  author={Anna Manasyan and Maximilian Seitzer and Filip Radovic and Georg Martius and Andrii Zadaianchuk},
  booktitle={Proceedings of the IEEE/CVF Conference on Computer Vision and Pattern Recognition (CVPR)},
  pages={5401--5411},
  year={2025},
}

\end{document}